\pdfoutput=1
\PassOptionsToPackage{table}{xcolor}
\documentclass[11pt]{article}

\usepackage[preprint]{acl}

\usepackage{times}
\usepackage{latexsym}

\usepackage[T1]{fontenc}
\usepackage[utf8]{inputenc}

\usepackage{microtype}

\usepackage{inconsolata}

\usepackage{graphicx}
\usepackage{subcaption}
\usepackage{enumitem}
\usepackage{url}
\usepackage{amsmath}
\usepackage{multirow}
\usepackage{multicol}
\usepackage{array}
\usepackage{booktabs}
\usepackage[table]{xcolor}

\usepackage{algorithm}
\usepackage{algpseudocode}
\algrenewcommand\algorithmicrequire{\textbf{Input:}}
\algrenewcommand\algorithmicensure{\textbf{Output:}}

\newcommand{\zblue}[1]{\text{\textcolor{blue}{#1}}}

\definecolor{mygray}{gray}{0.8}

\title{Attention Entropy is a Key Factor: An Analysis of Parallel Context Encoding with Full-attention-based Pre-trained Language Models}

\author{
Zhisong Zhang$^{\dag}$, Yan Wang, Xinting Huang, Tianqing Fang\\
\textbf{Hongming Zhang, Chenlong Deng, Shuaiyi Li, Dong Yu}\\
Tencent AI Lab\\
$^{\dag}$\texttt{zhisonzhang@tencent.com}
}

\begin{document}
\maketitle
\begin{abstract}
Large language models have shown remarkable performance across a wide range of language tasks, owing to their exceptional capabilities in context modeling. The most commonly used method of context modeling is full self-attention, as seen in standard decoder-only Transformers. Although powerful, this method can be inefficient for long sequences and may overlook inherent input structures. To address these problems, an alternative approach is parallel context encoding, which splits the context into sub-pieces and encodes them parallelly. Because parallel patterns are not encountered during training, naively applying parallel encoding leads to performance degradation. However, the underlying reasons and potential mitigations are unclear. In this work, we provide a detailed analysis of this issue and identify that unusually high attention entropy can be a key factor. Furthermore, we adopt two straightforward methods to reduce attention entropy by incorporating attention sinks and selective mechanisms. Experiments on various tasks reveal that these methods effectively lower irregular attention entropy and narrow performance gaps. We hope this study can illuminate ways to enhance context modeling mechanisms.
\end{abstract}

\section{Introduction}

% context modeling is crucial
Large language models (LLMs) have demonstrated exceptional capabilities across various language tasks \citep{achiam2023gpt,dubey2024llama}. A key factor contributing to this success is their remarkable ability of context modeling. This capability forms the basics of instruction following \citep{ouyang2022training,bai2022training} and in-context learning \citep[ICL;][]{brown2020language,dong-etal-2024-survey}, enabling LLMs to comprehend contexts effectively. Consequently, LLMs can solve tasks directly when provided with appropriate prompts \citep{liu2023pre}.

\begin{figure}[t]
	\centering
	\begin{subfigure}[b]{0.49\textwidth}
		\includegraphics[width=0.975\textwidth]{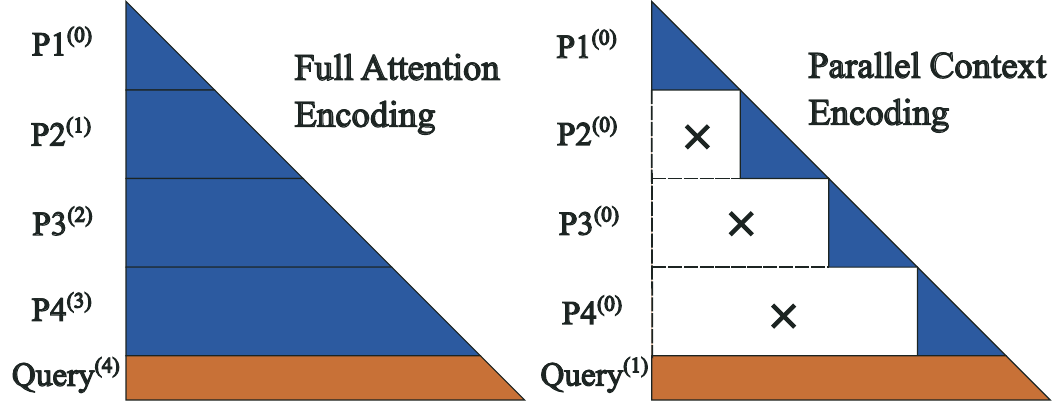}
	\end{subfigure}
	\begin{subfigure}[b]{0.49\textwidth}
		\includegraphics[width=0.975\textwidth]{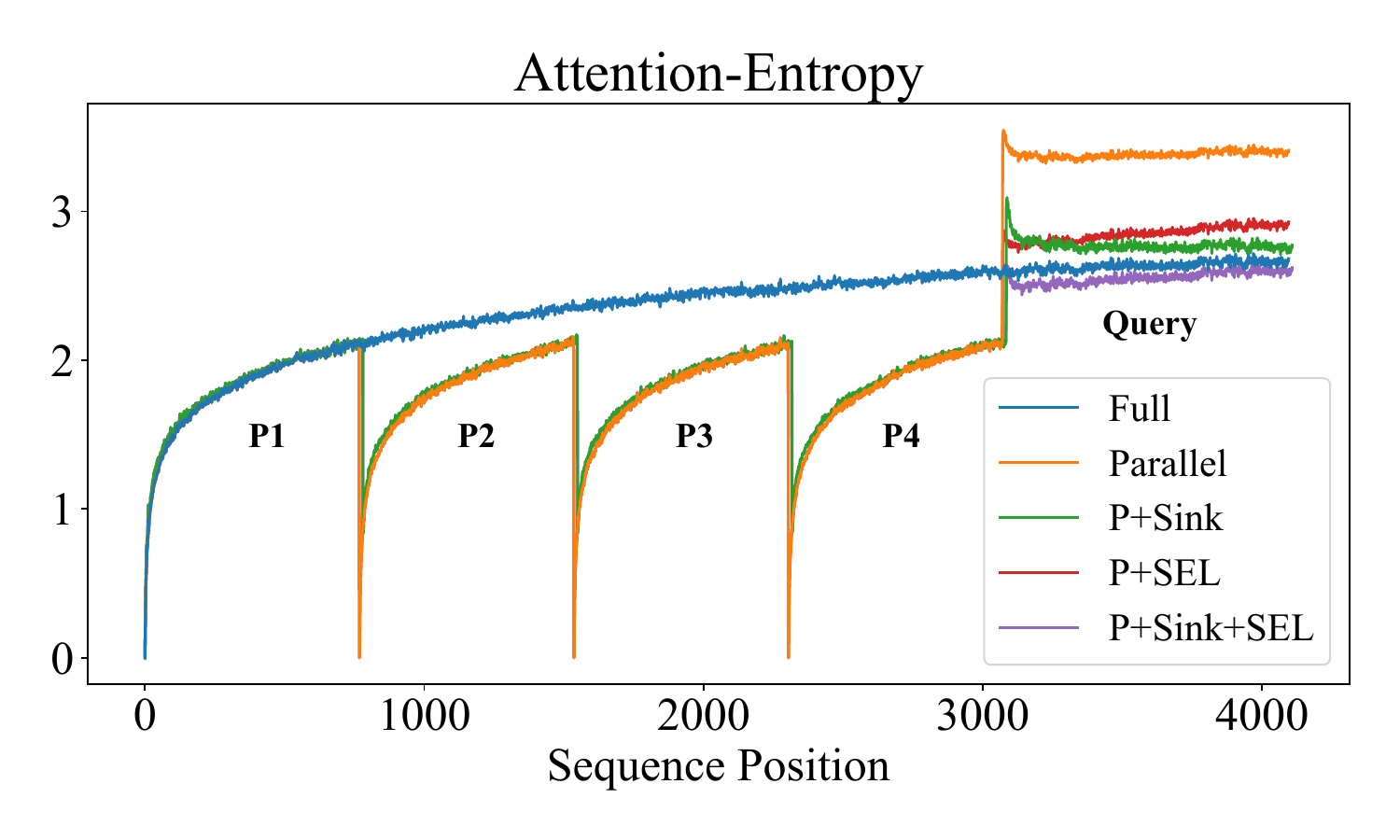}
	\end{subfigure}
	\caption{An overview. The upper sub-figures illustrate the full attention and parallel context encoding schemes, with superscripts indicating positional encoding. The lower sub-figure demonstrates that parallel encoding can result in irregularly high attention entropy for query tokens. In this example, the context is divided into four segments (P1–P4), which are individually encoded with the parallel encoding scheme (upper right). These pieces correspond to the sub-regions in the lower sub-figure. We explore two methods to reduce entropy (\S\ref{sec:reduce}): ``Sink'' means adding attention sinks, and ``SEL'' means selective attention.}
	\label{fig:main}
    \vspace*{-3mm}
\end{figure}

% most LLM are based on full attention, but 1) efficiency, 2) natural structures
To model contexts, most LLMs adopt a similar architectural design: an auto-regressive decoder-only Transformer with full self-attention \citep{vaswani2017attention,radford2019language}. This architecture does not assume contextual independence, allowing each token to attend to all previous tokens. While powerful and flexible, this design is not without concerns. First, full attention requires computational complexity that scales quadratically with the input sequence length. This poses challenges for long sequence processing and necessitates more efficient alternatives \citep{tay2022efficient}. Additionally, in many applications, contexts or prompts exhibit \emph{natural parallel structures}, consisting of independent sub-pieces, such as documents in retrieval-augmented generation \citep[RAG;][]{lewis2020retrieval} and demonstrations in ICL. It is intuitive to leverage these structures more effectively.

% parallel context encoding
To enhance the efficiency of context encoding and leverage the input structures, a natural strategy is to divide the context into sub-pieces, encode each one in parallel, and then concatenate them for final use. Figure~\ref{fig:main} illustrates the differences between full attention encoding and parallel context encoding. Compared to full encoding, the parallel approach can reduce the computational complexity since each sub-piece does not interact with others during the context encoding phase, and the parallel input structures are directly utilized for context splitting. However, mainstream LMs typically rely on full attention and haven't been trained with parallel contexts, posing the question of \emph{whether parallel context encoding is compatible with full-attention-based pre-trained LMs}. While specialized fine-tuning can ensure compatibility, it can also be computationally costly \citep{yen-etal-2024-long,sun2024block,lu2024turborag}. Recent studies have explored settings that do not require additional fine-tuning but these are limited to specific scenarios, such as restricted numbers of context windows \citep{ratner-etal-2023-parallel} or specific tasks like ICL \citep{hao2022structured} or RAG \citep{merth2024superposition}. In contrast, through detailed evaluations over various tasks, we provide more comprehensive analyses of this question, showing the connections between irregular attention entropy and the final performance.
% Such parallel context encoding scheme has been studied in recent work \citep{ratner-etal-2023-parallel,yen-etal-2024-long,hao2022structured,merth2024superposition,sun2024block,lu2024turborag}. However, most of these studies either require a costly fine-tuning procedure \citep{yen-etal-2024-long} or focus on specific scenarios such as limited context window numbers \citep{ratner-etal-2023-parallel} or specific downstream tasks like ICL \citep{hao2022structured} or RAG \citep{merth2024superposition,sun2024block,lu2024turborag}. In this work, we directly apply parallel context encoding to pre-trained LLMs on various tasks, aiming to answer the following research question:\\
% \emph{$\phantom{ab}$Is parallel context encoding compatible with full-attention pre-trained LLMs without requiring additional fine-tuning procedures?}

% what we explore and what we find
We conduct experiments on a variety of language tasks, including language modeling (LM), ICL, RAG and synthetic tasks. Through fair and direct comparisons between full-attention and parallel encoding schemes, we demonstrate that naively applying parallel encoding results in significant performance declines. By analyzing the attention patterns of both schemes, we find that parallel encoding leads to higher attention entropy on the final query tokens (Figure~\ref{fig:main} shows a typical example). Furthermore, we discover strong correlations between attention entropy and model performance, suggesting that attention entropy can be an indicator of irregular performance. To address this, we adopt two straightforward methods to reduce attention entropy: \emph{attention sinks} \citep{xiao2024efficient}, which adds a shared prefix to the context that each sub-piece can attend to, and \emph{selective attention}, which incorporates a hard selection mechanism into the attention operation. Experimental results show that both methods can reduce the irregular attention entropy and mitigate the performance gaps, verifying our hypothesis. Additionally, we provide a detailed analysis of how different selective attention choices affect performance across various tasks. We hope that our analysis could offer insights into exploring alternative context-modeling mechanisms beyond the full attention scheme.

\section{Preliminaries and Settings}

\subsection{Attention Entropy}

One central metric in our analysis is \textbf{attention entropy}, defined as $H(p) = -\sum_i p_i \cdot \log p_i$, where $p$ represents the attention probability distribution (attention weights). This metric quantifies the diversity of an attention head's focus over previous contexts. A higher entropy value indicates that the query token distributes its attention broadly across many prior tokens, suggesting a more global contextual understanding. On the other hand, a lower entropy value reflects a more concentrated attention pattern, which is crucial for tasks requiring precise retrieval.

\subsection{Parallel Context Encoding}
\label{sec:pce}

\begin{table*}[t]
	\centering
	\small
	\begin{tabular}{l | c c c | c c c | c c c | c c c}
		\toprule
		  & \multicolumn{3}{c|}{LM (PPL$\downarrow$)} & \multicolumn{3}{c|}{ICL (Acc$\uparrow$)} & \multicolumn{3}{c|}{RAG (SubEM$\uparrow$)} & \multicolumn{3}{c}{Synthetic (SubEM$\uparrow$)} \\
        \cmidrule(r){2-4} \cmidrule(r){5-7} \cmidrule(r){8-10} \cmidrule(r){11-13}
		& 4K & 8K & 16K & 4K & 8K & 16K & 4K & 8K & 16K & 4K & 8K & 16K \\
		\midrule
        Full & 5.54 & 5.35 & 5.19 & 55.20 & 66.00 & 72.20 & 61.25 & 60.25 & 60.25 & 99.88 & 99.50 & 97.25 \\
        \midrule
        P=2 & 5.83 & 5.66 & 5.47 & 50.80 & 63.60 & 70.20 & 61.50 & 59.50 & 57.50 & 93.81 & 94.81 & 95.25 \\
        P=4 & 6.29 & 6.16 & 6.04 & 36.40 & 57.20 & 67.80 & 59.25 & 50.75 & 52.50 & 79.19 & 81.56 & 82.44 \\
        P=8 & 6.91 & 6.92 & 6.96 & 29.20 & 44.40 & 60.20 & 53.50 & 48.75 & 44.50 & 25.94 & 41.00 & 41.44 \\
        \rowcolor{mygray}
        P=16 & 7.69 & 7.97 & 8.54 & 21.00 & 34.00 & 46.40 & 49.00 & 41.75 & 40.00 & 3.38 & 2.19 & 2.00 \\
        \rowcolor{mygray}
        P=32 & 8.54 & 9.24 & 10.87 & 10.80 & 17.40 & 33.60 & 45.00 & 39.25 & 35.25 & 0.31 & 0.00 & 0.00 \\
        \rowcolor{mygray}
        P=64 & 9.35 & 10.46 & 13.18 & 5.00 & 10.80 & 19.80 & 45.00 & 33.00 & 26.75 & 0.00 & 0.00 & 0.00 \\
		\bottomrule
	\end{tabular}
	\caption{Comparisons between full-attention and naive parallel encoding with \textsc{Llama-3.1-8B} (results are macro-averaged over all sub-tasks). Here, ``P'' indicates the parallel degree (the number of context sub-pieces). For each task, we also vary the total sequence lengths (considering 4K, 8K and 16K). With larger ``P'' (larger than 10), the results are \colorbox{mygray}{much worse} than those of full attention.}
	\label{tab:main_res}
\end{table*}

In a vanilla Transformer-based LM, to encode a context sequence of $N$ tokens (assuming the use of a decoder-only model with causal masks), each token needs to attend to all preceding tokens in the context. Consequently, we need to calculate the attention scores for $\frac{1}{2}\cdot{}N(N+1)$ token pairs. With parallel context encoding, we split the context into $P$ sub-pieces.\footnote{We refer to the number of sub-pieces as \emph{parallel degree}, which is one of the main variables examined in this study.} In this scheme, each piece is encoded separately, and tokens within one piece do not attend to tokens in other pieces. Assuming that we evenly split the context for simplicity, the token-pair calculation requirement is $P\cdot{}\frac{1}{2}\cdot{}\frac{N}{P}(\frac{N}{P}+1)=\frac{1}{2P}\cdot{}N(N+P)$, which is approximately $\frac{1}{P}$ of the computations needed in full attention. Therefore, the more pieces the context is split into, the more computational savings can be achieved. 
% Nevertheless, there is an efficiency-performance trade-off here: as will be shown in later experiments (\S\ref{sec:entropy}), more splitting introduces more irregularities and may lead to worse performance.

The parallel scheme is intuitive in many applications, such as RAG and ICL. This is because each piece, such as a document in RAG or a demonstration in ICL, is self-contained and does not require additional information in its encoding phase. The main phase where we need to check full contexts and aggregate information across pieces is the query-encoding phase. At this stage, we can let the querying tokens attend to the all preceding tokens to gather information.

While there can be minor variations, the basic methodology for parallel context encoding remains largely the same in previous research. Following \citet{ratner-etal-2023-parallel}, we provide a brief introduction of the two main modifications to full attention: \emph{position encoding} and \emph{attention masking}.

For \emph{position encoding}, each piece is parallel to each other and uses its own position counting mechanism. If the pieces have different lengths, we take the maximum length as the target context length and evenly distribute the position encoding of the tokens within each piece accordingly.\footnote{We explore models that utilize RoPE \citep{su2024roformer}, which allows for the assignment of real-valued position IDs. There can be other options for position encoding, including using the harmonic mean as the target length \citep{merth2024superposition} or retaining natural integer counting \citep{ratner-etal-2023-parallel}. Our choice is based on its overall good performance in preliminary experiments with our settings.} For example, assume we have three context pieces with lengths of $L1$, $L2$, and $L3$. With full attention, we need to arrange them sequentially and assign positions ranging from $0$ to $L1+L2+L3$ to all the tokens. With parallel encoding, we no longer need a specific order among different pieces; each piece independently counts its own tokens' positions starting from $0$ to the target length.

For \emph{attention masking}, we design special attention masks in accordance with the parallel encoding scheme. Each token within a context piece is restricted to attend only to the preceding tokens within this piece but not to other pieces. However, the final query tokens can attend to all preceding tokens across all context pieces to aggregate information. This approach results in inherently sparse attention calculations, for which sparse attention tools, such as FlexAttention\footnote{\url{https://pytorch.org/blog/flexattention/}} and block-sparse attention in FlashAttention \citep{dao2022flashattention}, can be used to enhance efficiency. 
% In this work, we primarily focus on performance analysis and leave efficiency analysis to future work.

\begin{figure*}[t]
	\centering
	\begin{subfigure}[b]{0.45\textwidth}
		\includegraphics[width=0.975\textwidth]{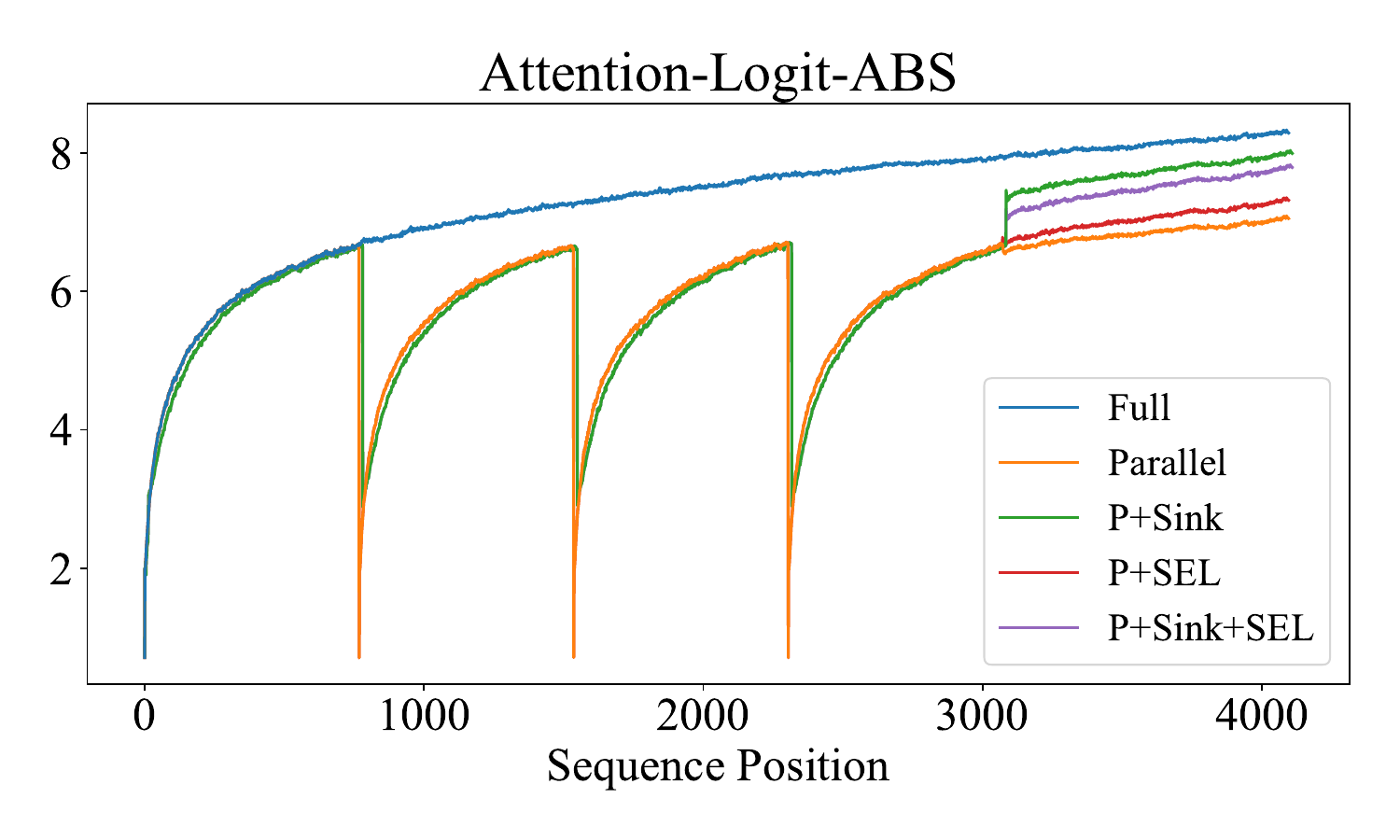}
	\end{subfigure}
    \begin{subfigure}[b]{0.45\textwidth}
		\includegraphics[width=0.975\textwidth]{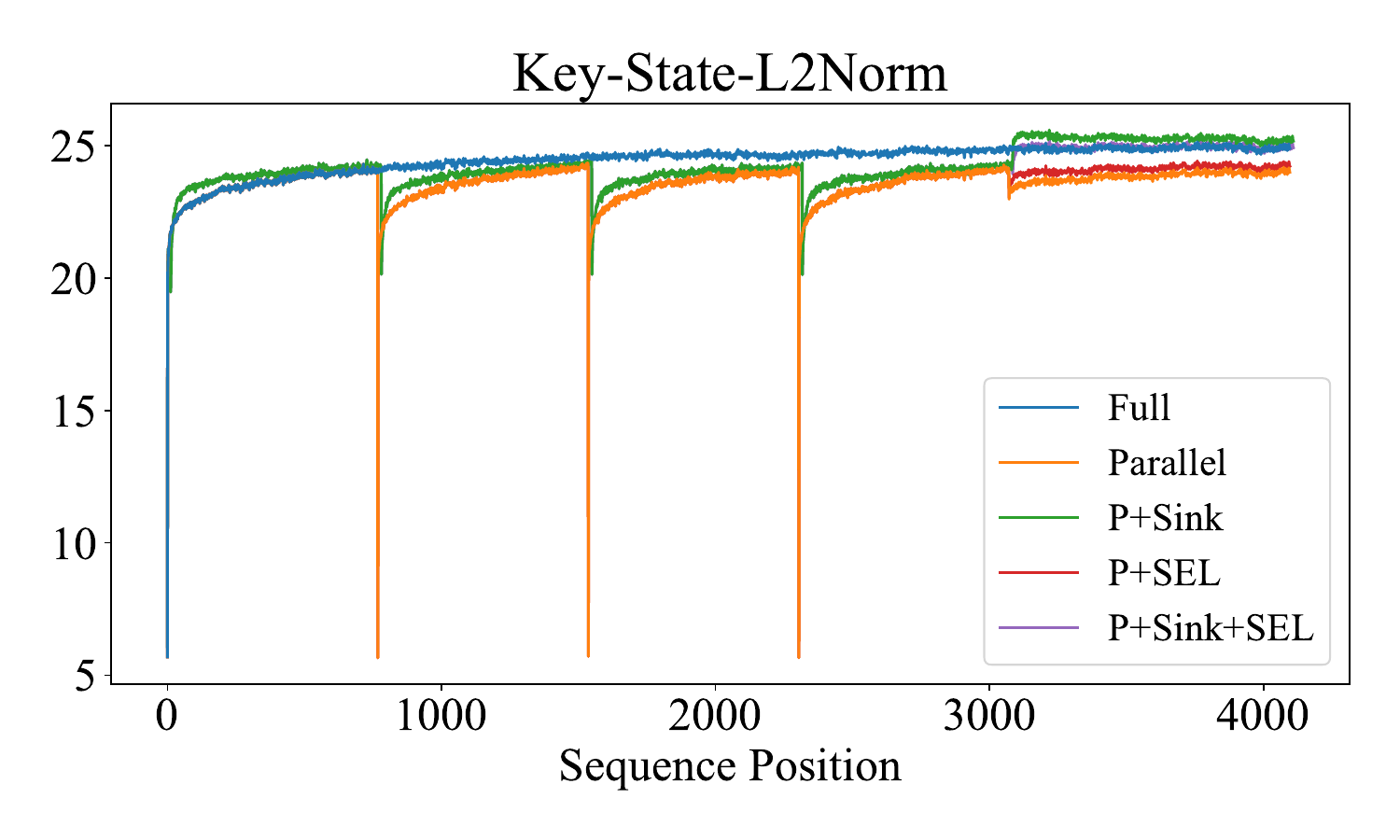}
	\end{subfigure}
	\caption{The scales of attention logits (averaged absolute values) and key states (L2 norm) with different methods. The irregularities of these scales may explain why attention entropy is higher with parallel context encoding.}
	\label{fig:ana}
    \vspace*{-3mm}
\end{figure*}

\subsection{Setups}

\paragraph{Task.} We experiment with a variety of language tasks to evaluate the influence of parallel context encoding, including LM, ICL, RAG and synthetic recall tasks. For LM, we use the PG19 \citep{Rae2020Compressive} and Proof-Pile \citep{azerbayev2023proofpile} datasets for evaluation. For the remaining tasks, we take the corresponding datasets from the HELMET benchmark \citep{yen2024helmet} and follow its processing protocols. Specifically, these include TREC-coarse and TREC-fine \citep{li-roth-2002-learning}, BANKING77 \citep{casanueva-etal-2020-efficient}, CLINC150 \citep{larson-etal-2019-evaluation} and NLU \citep{Liu2019BenchmarkingNL} for ICL;  Natural Question \citep{kwiatkowski-etal-2019-natural}, TriviaQA \citep{joshi-etal-2017-triviaqa}, HotpotQA \citep{yang-etal-2018-hotpotqa} and PopQA \citep{mallen-etal-2023-trust} for RAG; and three typical needle-in-a-haystack tasks \citep{kamradt2023needle} from RULER \citep{hsieh2024ruler} as well as a JSON retrieval task \citep{liu-etal-2024-lost} for synthetic recall.

\paragraph{Evaluation.} For all tasks, we assume that an input instance consists of a context and a query. The context can be further split into sub-pieces, for which we can apply parallel encoding, and the query can always attend to all previous contexts. For non-LM tasks, this scheme is natural: each instance already contains a query and a context consisting of a collection of items (documents in RAG, demonstrations in ICL, and haystack items in synthetic recall). Note that for parallel encoding, we can group multiple items into one sub-piece when we want a parallel degree (i.e., the number of parallel sub-pieces) that is smaller than the number of available items. For LM, we simulate this scheme by designating the final 1K tokens in a text segment as the query. The preceding tokens are considered as the context and are evenly divided into sub-pieces for parallel encoding. To evaluate the performance, we measure the perplexity (PPL) of the query tokens for LM. For other tasks, we follow HELMET and measure substring exact match for RAG and synthetic recall, and accuracy for ICL by comparing the model's generated output (with greedy decoding) to the gold answers.

\paragraph{Model.} 
We use \textsc{Llama-3.1-8B} as the primary model in our main experiments. Results from other models, including the \textsc{Instruct} version, \textsc{Mistral-7B-v0.3} and \textsc{Qwen2-7B}, exhibit similar overall trends and are detailed in Appendix \ref{sec:extra_res}. These models share a similar architecture design with RoPE-based positional encoding, which is adaptable and facilitates modifications for parallel encoding. Our experiments utilize the pre-trained models as they are, without any fine-tuning. However, modifications to the internal attention layers are necessary, which is why we cannot evaluate closed-source LLMs.

\section{Attention Entropy as an Indicator of Irregularities}
\label{sec:entropy}

\begin{figure}[t]
	\centering
	\begin{subfigure}[b]{0.45\textwidth}
		\includegraphics[width=0.975\textwidth]{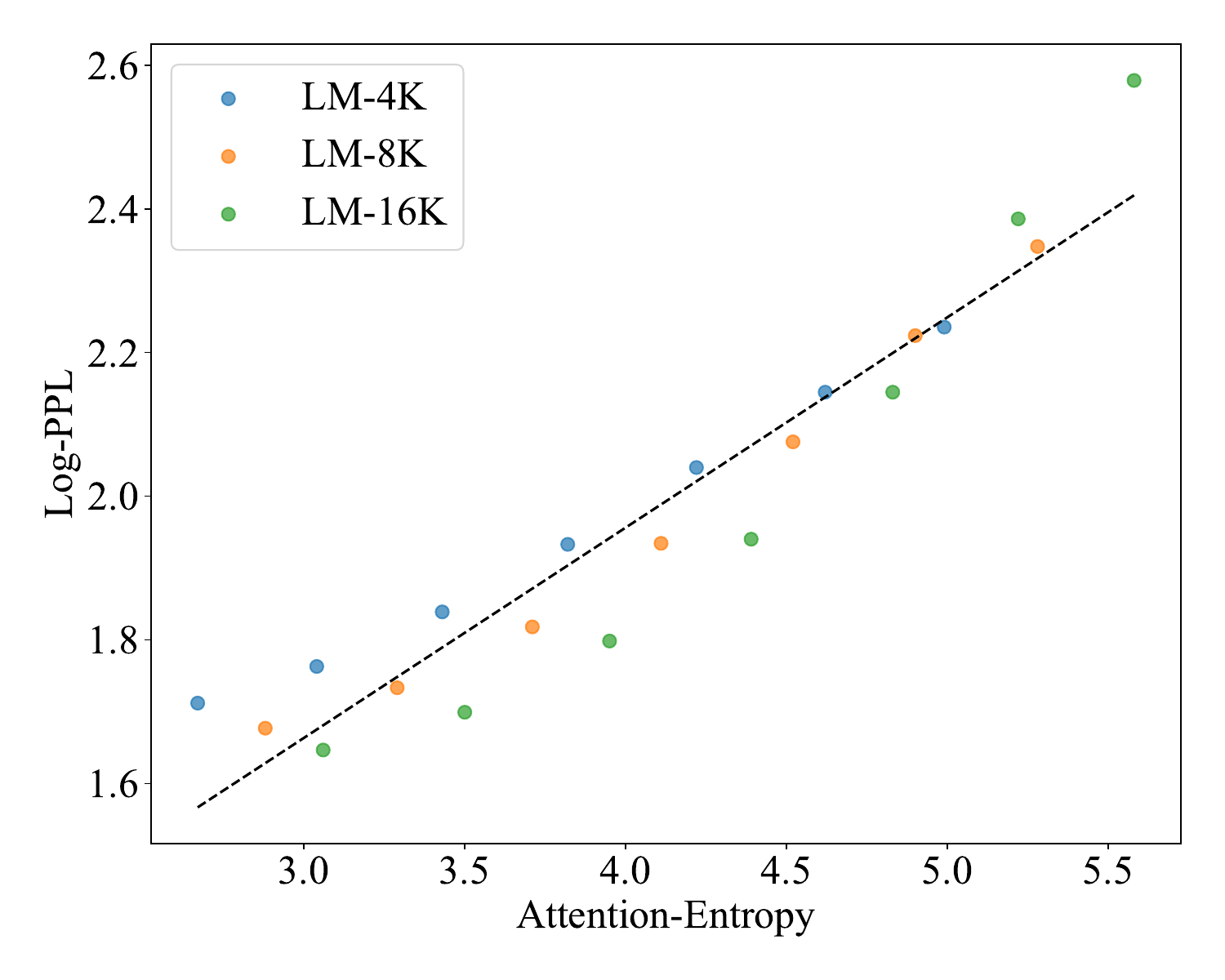}
	\end{subfigure}
	\caption{An illustration of the correlations between model performance and attention entropy (with the LM task and \textsc{Llama-3.1-8B}).}
	\label{fig:corr}
    \vspace*{-3mm}
\end{figure}

\paragraph{Naively applying parallel encoding leads to poorer performance.}
Table~\ref{tab:main_res} presents the main results of directly applying parallel context encoding to different tasks. Across all tasks, the direct application leads to worse results, and the performance degradation becomes more pronounced as the parallel degree increases. Notably, we can observe a dramatic decline for synthetic recall tasks: from nearly perfect accuracy with full attention to nearly complete failure when the context is split into tens of sub-pieces. This outcome is somewhat expected, as LLMs are trained with full attention and are thus unaccustomed to parallel contexts. This suggests that there could be some irregularities in the internal states of LLMs that are likely causing this failure.

\paragraph{Attention entropy can be an indicator of irregularities.}
Inspired by recent studies in LLM length extrapolation \citep{han-etal-2024-lm}, we examine and compare the attention values of different context encoding schemes. In length extrapolation, it is intuitive that longer sequences result in higher attention entropy values. Interestingly, we also observe irregularly high attention entropy for the query tokens when attending to parallel contexts. The lower part of Figure~\ref{fig:main} shows a typical example: it shows the averaged attention entropy values for the PG19 LM task (4K) with \textsc{Llama-3.1-8B}\footnote{Results are averaged over all layers and heads. Results with other models and tasks show similar patterns.} and a parallel degree of four. It demonstrates that when attending to parallel contexts, the attention entropy is much higher than that with vanilla full attention. Higher entropy usually denotes a higher level of uncertainty and confusion, which might explain why LLMs struggle to accurately retrieve information from the parallelly encoded contexts. Figure~\ref{fig:corr} illustrates the relationship between attention entropy and model performance, revealing strong correlations (\textsc{PearsonR}$\approx$0.95) between them. 
% This suggests that attention entropy can be a key indicator of the irregularities of parallel context encoding.

\paragraph{Irregularly high entropy can be attributed to irregularities in hidden state scales.}
We further investigate\footnote{For this analysis, we again average over all the layers and heads. Note that, for these scales, we observe larger variations among different heads than those in the entropy analysis. Nevertheless, we think that the averaged results can still meaningfully provide an overall explanation.} the causes of irregularly high attention entropy. Firstly, we examine the scales of attention logits -- the input to the attention softmax operation. As shown in the left sub-figure of Figure~\ref{fig:ana}, we again find irregularities: the averaged absolute values of attention logits are smaller with parallel encoding. To examine what causes the irregular logit scales, we further inspect the key states -- the inputs to the \textsc{MatMul} operation that produces the attention logits. As illustrated in the right sub-figure of Figure~\ref{fig:ana}, the norm of the key states generally increases along the sequence dimension. With parallel context encoding, where the context pieces are encoded individually, the key states have smaller norms than those in full attention. Especially, the initial tokens in each piece, which are known as sink tokens \citep{xiao2024efficient}, have dramatically smaller norms \citep{gu2024attention}. While it would be interesting to further investigate the cause of the irregular hidden state patterns, we find that this involve complex interactions with various Transformer layers, such as LayerNorm and MLP; a complete explanation of this phenomenon would require a deeper understanding of the underlying working mechanisms of Transformers, which we leave to future exploration.

\section{Reducing Entropy with Attention Sinks and Selective Attention}
\label{sec:reduce}

\begin{figure}[t]
	\centering
	\begin{subfigure}[b]{0.4\textwidth}
		\includegraphics[width=0.975\textwidth]{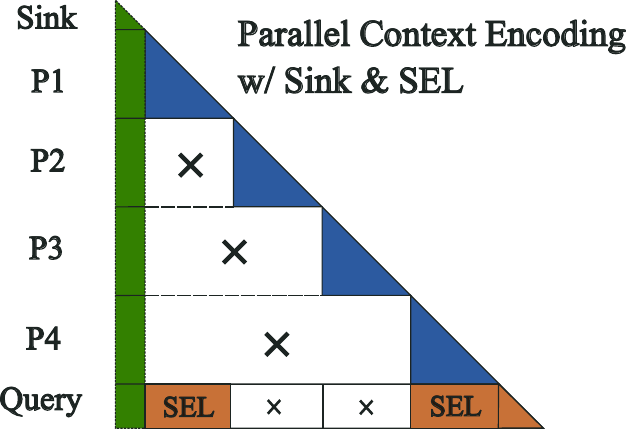}
	\end{subfigure}
	\caption{Illustrations of our methods to reduce attention entropy: adding shared attention sinks (Sink) and adopting selective attention (SEL).}
	\label{fig:reduce}
    \vspace*{-3mm}
\end{figure}

\subsection{Methods}

Our prior analysis indicates a strong correlation between model performance and attention entropy. However, \emph{correlation does not imply causation}. To investigate whether the irregular attention entropy is a key factor of performance degradation, we adopt two straightforward methods to adjust the attention entropy, attention sinks and selective attention, as depicted in Figure~\ref{fig:reduce}.

\subsubsection{Attention Sinks} 
Recent studies on attention sinks have demonstrated that initial tokens significantly influence the internal dynamics of Transformers \citep{xiao2024efficient, han-etal-2024-lm, gu2024attention}. As shown in Figure~\ref{fig:ana}, we also observe that the sinking tokens exhibit abnormal hidden state scales, potentially leading to irregular attention entropy. When naively applying parallel context encoding, each sub-piece contains its own sinks, which are subsequently attended to by later query tokens. The model has never encountered such multi-sink patterns in LM training and thus produces irregular hidden states. To mitigate this problem, we prepend a shared prefix to all the context sub-pieces to eliminate attention sinks inside each sub-piece. Interestingly, preliminary experiments indicate that the specific content of the shared prefix is not crucial; even adding tokens of linebreaks can be effective, indicating that their main functionality is to absorb unneeded attention values. Without loss of generality, we manually write simple instructions\footnote{For LM, we use \textit{``Given the following partial context, predict the next sequence of words:''}; for other tasks, we use \textit{``Given the following contexts, answer the final question accordingly:''}.} as the shared prefixes.

The impact of incorporating shared attention sinks can be examined by analyzing LM's internal states. As shown in Figure~\ref{fig:ana}, attention sinks can avoid the extremely irregular tokens in each sub-piece (the original sink tokens) and lead to higher attention logit values, which lead to lower attention entropy as depicted in Figure~\ref{fig:main}. As discussed in the following subsection, this can indeed enhance performance, suggesting that shared attention sinks can help the model to be more familiar with the hidden state patterns of parallel context encoding.

\subsubsection{Selective Attention} 
An alternative method to reduce attention entropy is to directly sharpen the attention distribution through hard selection. Specifically, we group the context tokens according to the splitting of the parallel sub-pieces. A sub-piece score is then calculate for each group, followed by a top-K selection process for each attention operation. For instance, in the scenario of four context pieces as depicted in Figure~\ref{fig:reduce}, we select the top-2 scored sub-pieces and exclude the remaining two from the attention calculation. As shown in Figure~\ref{fig:main}, this selective mechanism can directly reduce entropy.

% -----
\begin{algorithm}[t]
    \caption{Selective Attention.}
	\label{algo:sel}
	\begin{algorithmic}[1]
		\small
		\Require Original attention probability tensor $p_{in}$. 
		\Ensure Modified attention probability tensor $p_{out}$.
        % --
        \State $s_{group} ~\leftarrow~ \text{group\_key}(p_{in})$ \Comment{\zblue{Obtain grouped scores}}
        % \State $s_{aggr} ~\leftarrow~ \text{aggregate}(s_{group})$ \Comment{\zblue{Aggregation}}
        \State $i_{sel} ~\leftarrow~ \text{top\_k}(s_{group})$ \Comment{\zblue{Group selection}}
        \State $m ~\leftarrow~ \text{expand\_mask}(i_{sel})$ \Comment{\zblue{Expand SEL mask}}
        \State $p_{m} ~\leftarrow~ p_{in} \cdot m$ \Comment{\zblue{Apply mask}}
        \State $p_{out} ~\leftarrow~ p_{m} / p_{m}.sum(-1)$ \Comment{\zblue{Re-normalization}}
        \State \textbf{return} $p_{out}$
		% --
	\end{algorithmic}
\end{algorithm}
% -----

The overall procedure is outlined in Algorithm~\ref{algo:sel}. It can be easily understood by examining the shapes of the intermediate tensors.
\begin{itemize}[leftmargin=*]
    \vspace*{-2mm}
    \item \textbf{Input.} The input attention probability tensor $p_{in}$ has the shape of $[N_{layer}, N_{head}, L_{query}, L_{key}]$.
    \vspace*{-2mm}
    \item \textbf{Grouping.} We first compute the group score for each sub-piece along the ``Key'' dimension: each group has a piece of attention probabilities, which are reduced into one group score. We use the sum of the top-5 values\footnote{Preliminary experiments indicate that the results are not very sensitive to the number of top values used in this step.} as the reduction function, which is found to be better than using sum or average. Assuming there are $P$ sub-pieces, the final group score $s_{group}$ will have the shape of $[N_{layer}, N_{head}, L_{query}, P]$.
    \vspace*{-2mm}
    \item \textbf{Selection.} The selection is performed along the group dimension, where only the top-K\footnote{We choose K=2 as the default value since earlier results (see Table~\ref{tab:main_res}) demonstrate that using two parallel contexts does not significantly impact the outcomes.} scored groups are selected as valid. We obtain the selected group indexes $i_{sel}$, which has the shape of $[N_{layer}, N_{head}, L_{query}, K]$.
    \vspace*{-2mm}
    \item \textbf{Masking.} The selected indexes are expanded to obtain the selection mask over the original tokens. Tokens within parallel contexts that do not belong to any selected groups will be masked out. This mask $m$ has the same shape as $p_{in}$.
    \vspace*{-2mm}
    \item \textbf{Output.} Finally, the mask $m$ is applied to the input probability tensor (attention weights), and the final output attention probability tensor is obtained after a final re-normalization step to ensure that each row sums up to one.
\end{itemize}
Between the grouping and selection step, an optional reduction operation can be performed to aggregate information among tokens, heads or even layers. For example, if aggregating over the query-token dimension, $s_{group}$ is reduced from $[N_{layer}, N_{head}, L_{query}, P]$ to $[N_{layer}, N_{head}, 1, P]$. This reduction is useful in scenarios where the most relevant information comes from the same sub-piece for all tokens in the current query. If aggregating over heads, it can help to identify the most salient information-seeking head, such as the retrieval head \citep{wu2024retrieval}. A more aggressive reduction can be performed across the first three dimensions, reducing the group score to $[1, 1, 1, P]$, which is exactly the same as a retrieval procedure. Notice that if we choose the layer dimension for aggregation,\footnote{For other cases (without layer aggregation), this operation will bring only slight overhead, since we incrementally apply the selection mechanism for each layer (first performing softmax to obtain the attention scores, then performing selection, and finally re-normalizing).} we need to forward the model twice since attention scores at later layers are not available when calculating previous layers; for other dimensions, the selective attention modification can be applied immediately after each attention score is calculated. We again use the sum of top-5 values as the reduction function to identify the most salient attention scores.

We do not apply aggregation for the LM task since it often requires diverse information from their contexts; for other tasks where there are clear queries and information sources, we reduce on the head and query dimensions by default, which is found to perform well overall. We provide further analyses on the specifications of selection attention for different tasks in \S\ref{sec:var_sel}.

\subsection{Main Results}
\label{sec:results}

\begin{figure*}[t]
	\centering
	\begin{subfigure}[b]{0.95\textwidth}
		\includegraphics[width=0.975\textwidth]{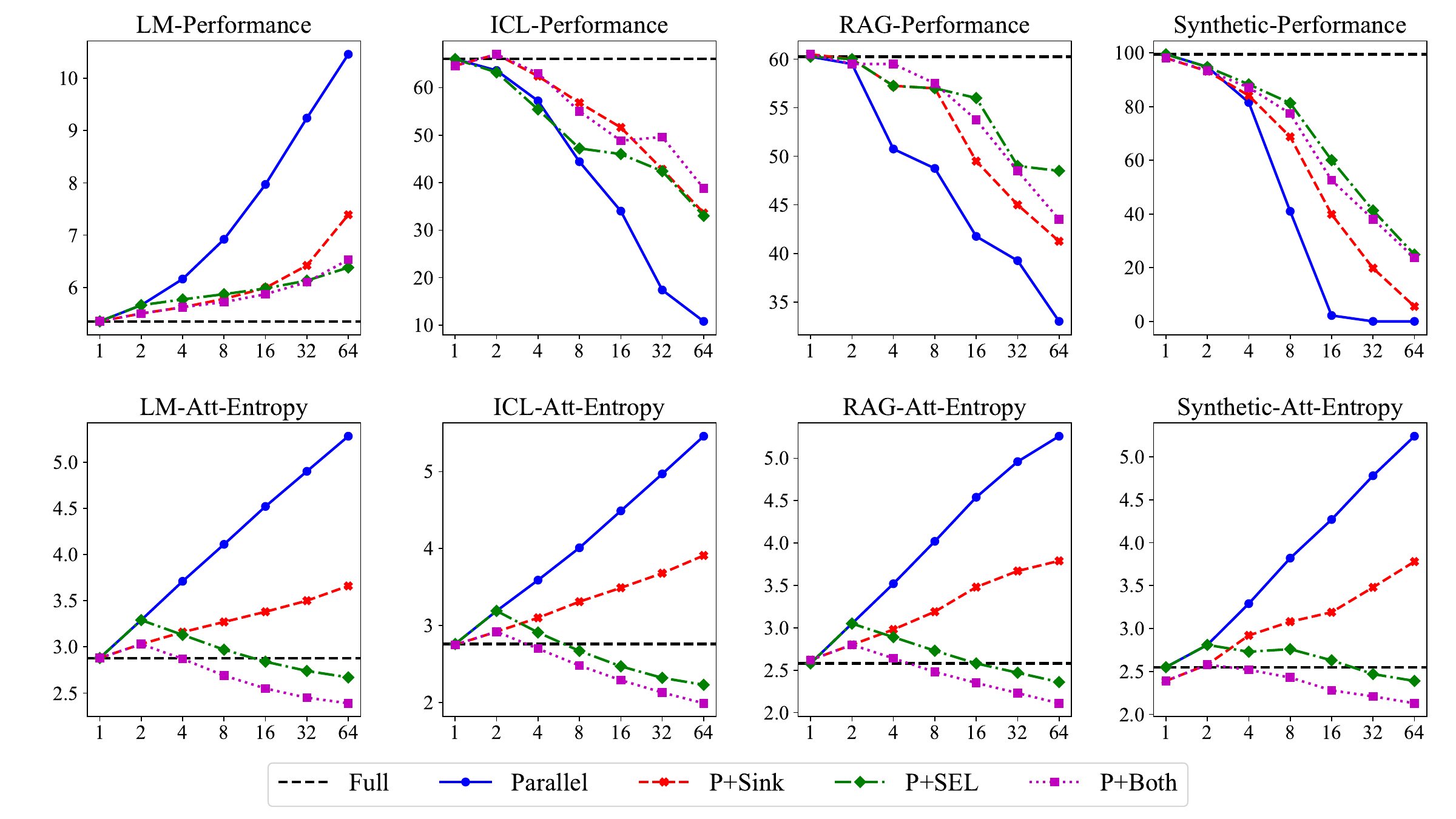}
	\end{subfigure}
	\caption{Results of entropy reduction methods (with \textsc{Llama-3.1-8B} and 8K lengths). The $x$-axis denotes the parallel degree $P$. The upper figures illustrate the model's performance: PPL for LM (the lower the better) and Accuracy or SubEM for other tasks. The lower figures denote the averaged attention entropy over the query tokens. Additional results for more models and settings are presented in Appendix~\ref{sec:extra_res}.}
	\label{fig:methods}
    \vspace*{-2mm}
\end{figure*}

Figure~\ref{fig:methods} illustrates the effectiveness of the entropy reduction methods (with \textsc{Llama-3.1-8B} and 8K lengths). The overall trends are consistent across different tasks. First, both shared sink tokens and selective attention can reduce attention entropy and enhance performance compared to the naive parallel scheme, especially with higher parallel degrees. Additionally, these two methods impact attention entropy differently: with sinks, the entropy is lower than the naive scheme, but still grows larger than that of full attention ($P$=1); with selective attention, the entropy decreases and can sometimes become even lower than that of full attention. Lastly, the benefits of these methods vary depending on the task. Selective attention is more helpful for RAG and synthetic recall tasks, which is intuitive because of the retrieval nature of these tasks.  On the other hand, attention sinks seem to be more beneficial for ICL tasks, since these tasks may need information from more demonstration examples than our default selective top-K value (K=2). Combining both techniques offers a balanced approach and yields overall effective performance.

\subsection{Analyses}

\subsubsection{Variations on Selection Attention}
\label{sec:var_sel}

We provide detailed ablation studies on various choices in the selection attention procedure. Since there are no clear query tokens in the LM task, our analysis primarily focuses on the other three tasks. We consider a typical scenario as the case study: using \textsc{Llama-3.1-8B} and 8K lengths, and adopting a difficult parallel degree of $P$=64. Firstly, we start with our default setting of aggregating over the Head and Token dimensions (denoted as ``HT''), and vary the K value in sub-piece top-K selection process. The results indicate that the optimal setting varies by task: synthetic recall tasks benefit from a small K value, since they require precise information from a few pieces, while ICL\footnote{Breaking down on ICL tasks, we further find some patterns: for example, for coarse-grained TREC (6 labels), the performance of ``K=1'' is closer to that of the best performing ``K=5'' (0.45 vs 0.48), while for other ICL tasks where there are tens of labels, ``K=1'' usually lags behind for more.} and RAG perform better with slightly larger K values, since additional context information can be helpful. Next, we examine different ways of information aggregation (with TopK=5 for ICL and RAG, and TopK=2 for synthetic tasks). Once again, different tasks exhibit distinct patterns: aggregating over all layers, as in a retrieval setting, yields the best results for RAG, layer-wise selection is more effective for synthetic tasks, and query-level selection is not crucial for ICL. While a universal and consistent method that performs well across all tasks would be ideal, achieving this may be difficult and costly. One advantage of our method is its flexibility, allowing dynamic adjustment of configurations to suit the specific nature of each task.

\begin{table}[t]
	\centering
	\small
	\begin{tabular}{l | c c c}
		\toprule
         & ICL & RAG & Synthetic \\
        \midrule
        TopK=1 & 26.00 & 45.75 & \underline{21.56} \\
        TopK=2 & \underline{33.00} & \underline{48.50} & \textbf{24.88} \\
        TopK=5 & \textbf{36.00} & \textbf{48.75} & 14.69 \\
        TopK=10 & 28.60 & 44.50 & \phantom{0}5.25\\
        \midrule
        No Aggr. & 35.40 & 42.75 & 17.62 \\
        Aggr.=T & \textbf{36.20} & 45.00 & \underline{21.00} \\
        Aggr.=HT & \underline{36.00} & \underline{48.75} & \textbf{24.88} \\
        Aggr.=LHT & 22.40 & \textbf{49.50} & 17.31 \\
		\bottomrule
	\end{tabular}
	\caption{Ablation studies on selection attention (with \textsc{Llama-3.1-8B}, 8K lengths and $P$=64). ``TopK'' denotes how many sub-pieces to select for each attention, ``Aggr.'' means the dimensions on which we apply aggregation (Layer, Head or Token).}
	\label{tab:sel_ana}
    \vspace*{-3mm}
\end{table}

\subsubsection{Effects of Value-only Parallel Encoding}
\label{sec:repl_key}

\begin{figure}[t]
	\centering
	\begin{subfigure}[b]{0.45\textwidth}
		\includegraphics[width=0.975\textwidth]{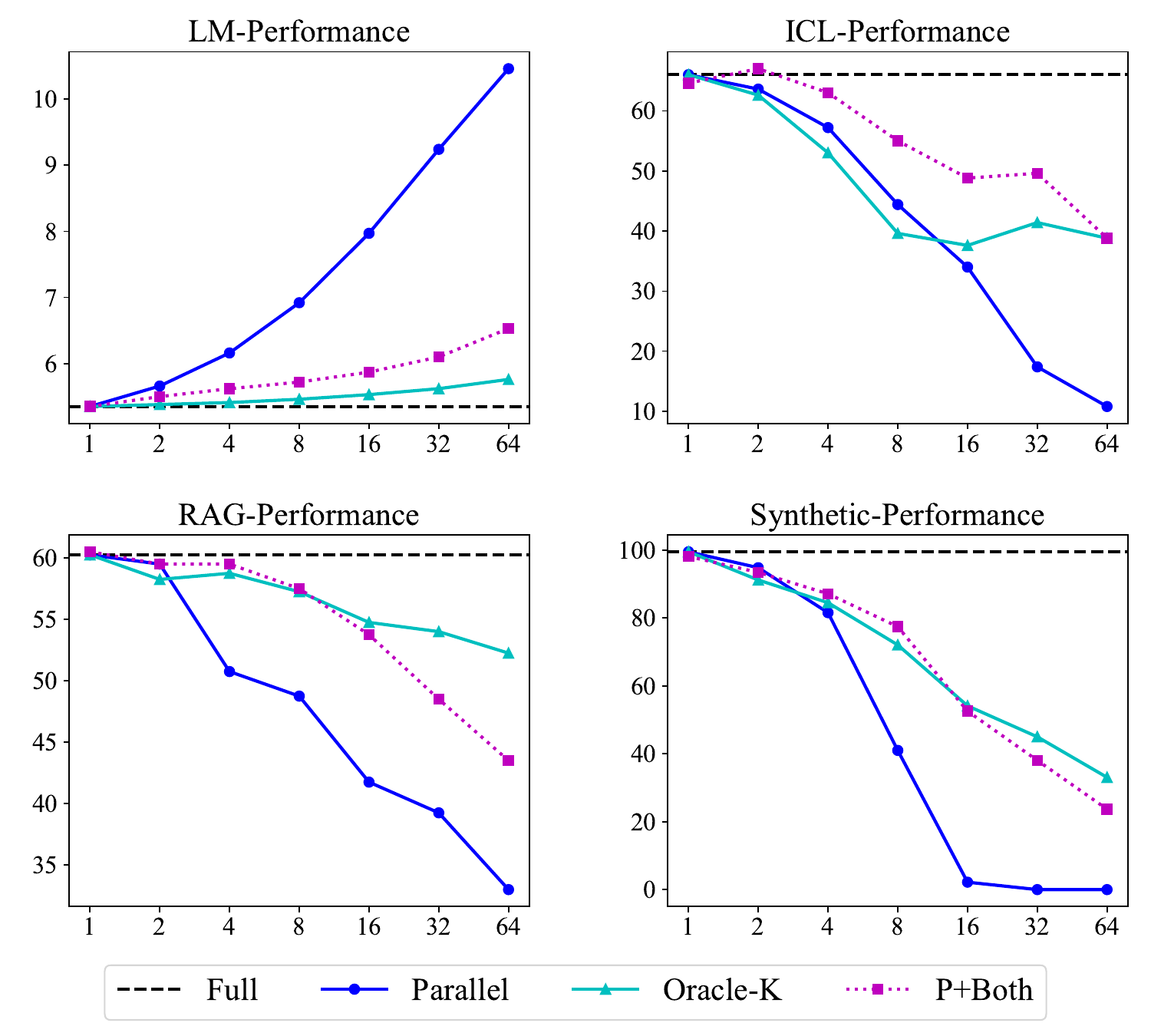}
	\end{subfigure}
	\caption{Performance with the oracle setting of value-only parallel encoding.}
	\label{fig:methods_kv}
\end{figure}

In our experiments, we mainly examine the attention patterns and methods to reduce attention entropy. In parallel context encoding, the value states are also influenced. To investigate the impact of value states, we consider an oracle setting\footnote{We've also tried only replacing value states, which leads to significantly worse and meaningless results.} where we replace the key states with those from full attention encoding; in this way, we have a mixed setting of value-only parallel encoding. Figure~\ref{fig:methods_kv} illustrates the results. Except for LM, using oracle key states does not always perform better than our methods, indicating that value states also play an important role in contextual encoding.

\section{Related Work}

\paragraph{Parallel Context Modeling.} Recent research has explored parallel context encoding for various tasks. \citet{ratner-etal-2023-parallel} present parallel context window to extend LLMs for handling longer contexts, which is beneficial for ICL and RAG tasks. Similarly, \citet{hao2022structured} scale ICL to accommodate thousands of demonstrations with a similar approach. \citet{yen-etal-2024-long} train an additional context encoder and cross-attention layers to achieve enhanced context encoding, albeit at a higher computational cost. Furthermore, parallel encoding has been applied to RAG \citep{merth2024superposition,sun2024block,lu2024turborag}, where the retrieved documents are naturally parallel to each other. Beyond encoding, the decoding process can be also made parallel, as explored in non-autoregressive generation \citep{stern2018blockwise,ghazvininejad-etal-2019-mask} and more efficient LLM prompting techniques \citep{ning2024skeletonofthought}.

\paragraph{Efficient Attention.} In addition to parallel context encoding, there has been considerable work on the topics of efficient attention \citep{tay2022efficient}. Adopting sparse attention patterns is a typical approach that selects certain tokens in the attention mechanism with either fixed \citep{child2019generating,beltagy2020longformer} or learned \citep{Kitaev2020Reformer,roy-etal-2021-efficient,gupta-etal-2021-memory} patterns. Parallel context encoding can be viewed as a special form of sparse attention, which enhances block sparsity. Another line of work focuses on efficient approximation of full attention using linear attention techniques \citep{katharopoulos2020transformers,choromanski2021rethinking,peng2021random}. Recently, with the advent of LLMs, there has been a renewed interest in efficient attention mechanisms for Transformer models to reduce computational and memory costs. Prompt or KV-cache compression techniques have been widely investigated, and the approaches mainly include training special compressing tokens \citep{mu2023learning,chevalier-etal-2023-adapting,ge2024incontext,qin-etal-2024-dodo,mohtashami2024random} or dynamically selecting tokens at inference time \citep{zhang2023h2o,liu2024scissorhands,ge2024model,li2024snapkv}. Our attention selection approach shares similar spirits to the latter strategies, though we perform the selection over context blocks.

\paragraph{Attention Analysis.} Since the introduction of self-attention in Transformers, analyzing the roles the attention mechanism plays has been a popular topic \citep{clark-etal-2019-bert,jain-wallace-2019-attention,serrano-smith-2019-attention,wiegreffe-pinter-2019-attention,bibal-etal-2022-attention}. The most relevant work to this study includes findings on attention sinks and specialized attention heads. Attention sinks refer to the initial tokens that attract most of the attention weights in many heads, and they have been utilized to extend LLMs to longer context lengths \citep{xiao2024efficient,han-etal-2024-lm,gu2024attention}. These works also inspire our analyses on attention entropy and hidden state norms. \citet{wang2024precision} also find that adding ``sink'' tokens improves performance from the perspective of precision. Additionally, it has also been shown that there can be specialized attention heads that perform special functions, such as syntactic heads for encoding syntactical relations \citep{clark-etal-2019-bert}, retrieval heads for collecting information from long contexts \citep{wu2024retrieval}, and induction heads that may constitute the mechanism for ICL \cite{olsson2022context}. Our reduction operations in selective attention are also based on the hypothesis that there is a small portion of information-seeking heads that can collect the most salient features from the contexts. Moreover, \citep{attanasio-etal-2022-entropy} propose Entropy-based Attention Regularization to mitigate bias by penalizing tokens with low self-attention entropy, demonstrating the usefulness of controlling attention entropy in a different scenario.

% no space for this right now
\section{Conclusion}
In this work, we present a detailed analysis of parallel context encoding for full-attention-based LMs without any fine-tuning. We demonstrate that naively applying parallel encoding leads to noticeably worse performance, particularly as the parallel degree increases. Through our analyses, we discover a strong correlation between irregularly high attention entropy and performance degradation. We adopt two approaches to reduce the entropy, which can help mitigate the performance gaps. We hope that our analyses and results can shed light on a deeper understanding and improvement of attention mechanisms.

% \newpage
\section*{Limitations}

This work has several limitations. First, we primarily use the pre-trained LM as it is without applying any fine-tuning. Clearly, fine-tuning could mitigate the irregularities in parallel encoding and enhance performance. However, it will bring extra computational costs, and selecting appropriate fine-tuning datasets would require careful consideration to maintain the model's general capabilities. Second, we mainly focus on performance analyses in this work, while leaving efficient implementation and related analyses to future work, which would require kernel-level implementations to achieve speed improvements. Finally, we have not found a universal and consistent method to fully address the performance gaps between full attention and parallel context encoding schemes. Further investigation and the incorporation with lightweight fine-tuning may help to close these gaps.

\section*{Ethics Statement}

This research primarily concentrates on analyses of language models. Consequently, we have not implemented any extra aggressive filtering techniques on the text data beyond the preprocessing done by the original dataset sources. We have also employed open-source language models in their existing form, without further addressing aspects such as enhancing safety and debiasing. As a result, the text data and models we used may contain issues related to offensiveness, toxicity, fairness, or bias that we have not identified, as these concerns are not the main focus of our study. Apart from these considerations, we do not foresee any additional ethical concerns or risks associated with our work.

\bibliography{main}

\newpage
\onecolumn
\appendix

\section{Additional Results}
\label{sec:extra_res}

\begin{figure}[h]
	\centering
	\begin{subfigure}[b]{0.325\textwidth}
		\includegraphics[width=0.975\textwidth]{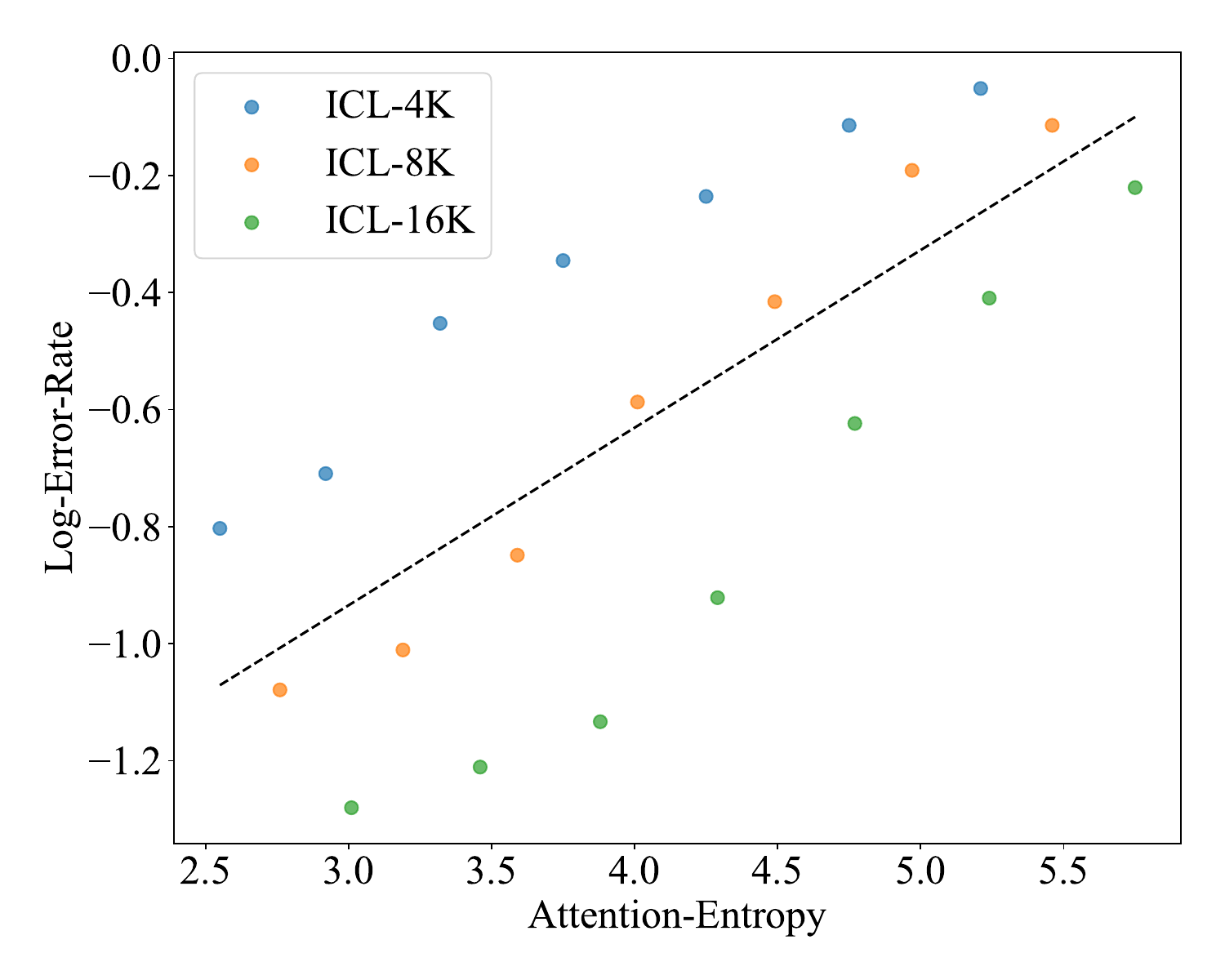}
	\end{subfigure}
    \begin{subfigure}[b]{0.325\textwidth}
		\includegraphics[width=0.975\textwidth]{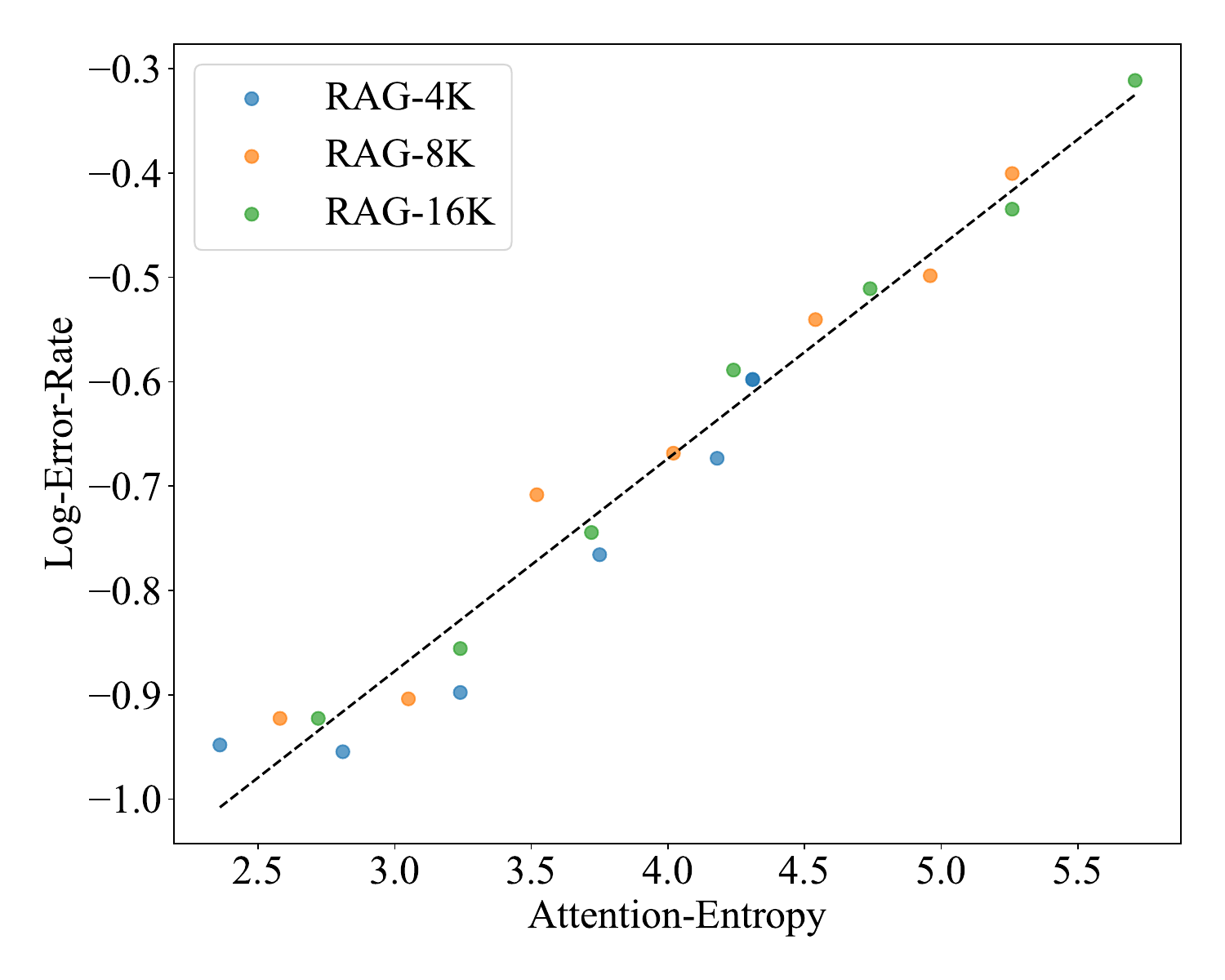}
	\end{subfigure}
    \begin{subfigure}[b]{0.325\textwidth}
		\includegraphics[width=0.975\textwidth]{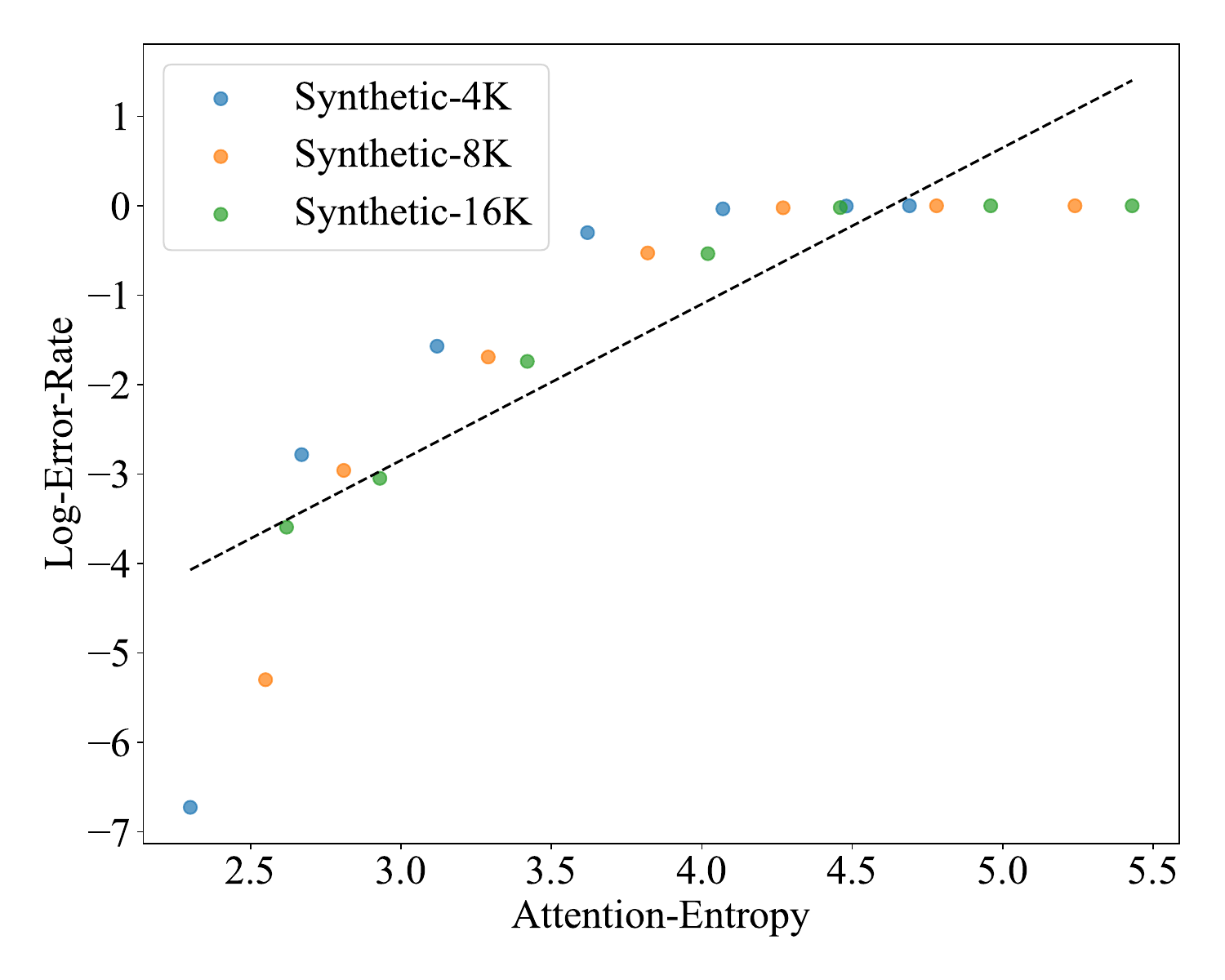}
	\end{subfigure}
	\caption{An illustration of the correlations between model performance and attention entropy on more tasks (with \textsc{Llama-3.1-8B}). Notations are similar to those in Figure~\ref{fig:corr}.}
	\label{fig:corr2}
\end{figure}

\begin{figure*}[t]
	\centering
	\begin{subfigure}[b]{0.95\textwidth}
		\includegraphics[width=0.975\textwidth]{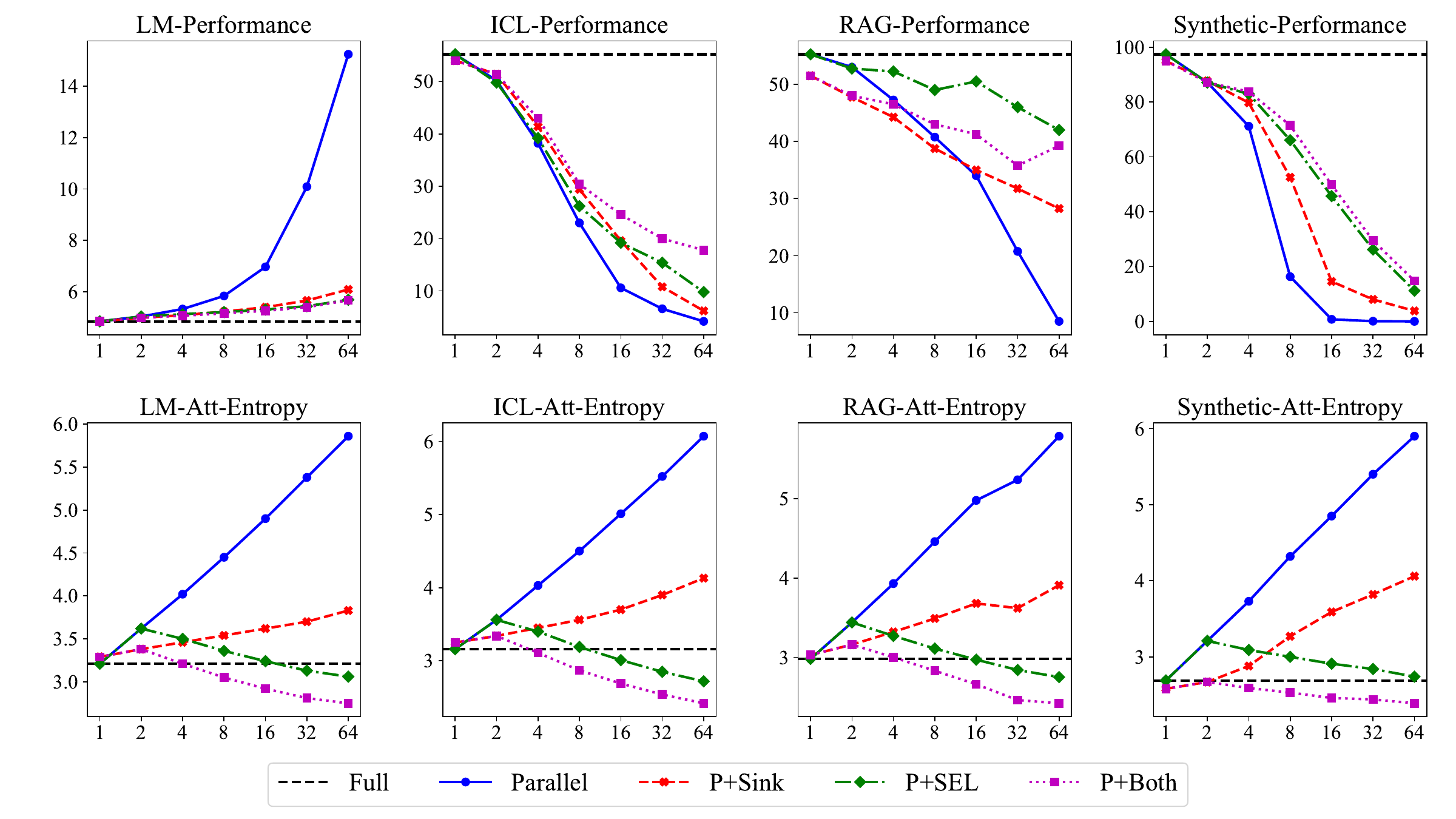}
	\end{subfigure}
	\caption{The influence of the entropy reduction methods (with \textsc{Mistral-7B-v0.3} and 8K lengths). Notations are the same as those in Table~\ref{fig:methods}.}
	\label{fig:methods2}
\end{figure*}

\begin{figure*}[t]
	\centering
	\begin{subfigure}[b]{0.95\textwidth}
		\includegraphics[width=0.975\textwidth]{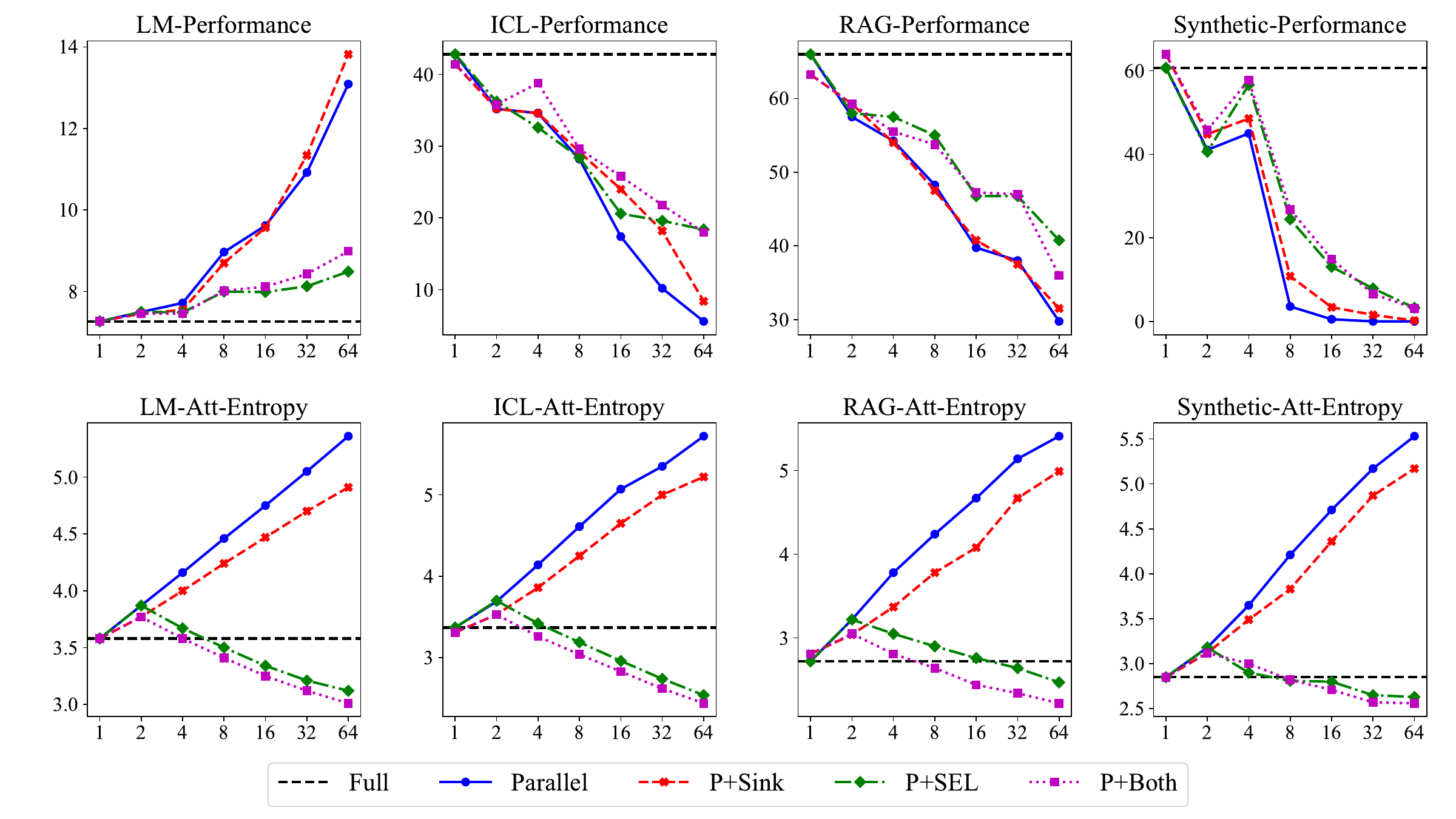}
	\end{subfigure}
	\caption{The influence of the entropy reduction methods (with \textsc{Qwen2-7B} and 8K lengths). Notations are the same as those in Table~\ref{fig:methods}. Note that the ``Sink'' method seems to be less effective for Qwen, probably because it is less influenced by sink tokens, as evidenced by the less entropy reduction brought by ``Sink''.}
	\label{fig:methods3}
\end{figure*}

\begin{figure*}[t]
	\centering
	\begin{subfigure}[b]{0.95\textwidth}
		\includegraphics[width=0.975\textwidth]{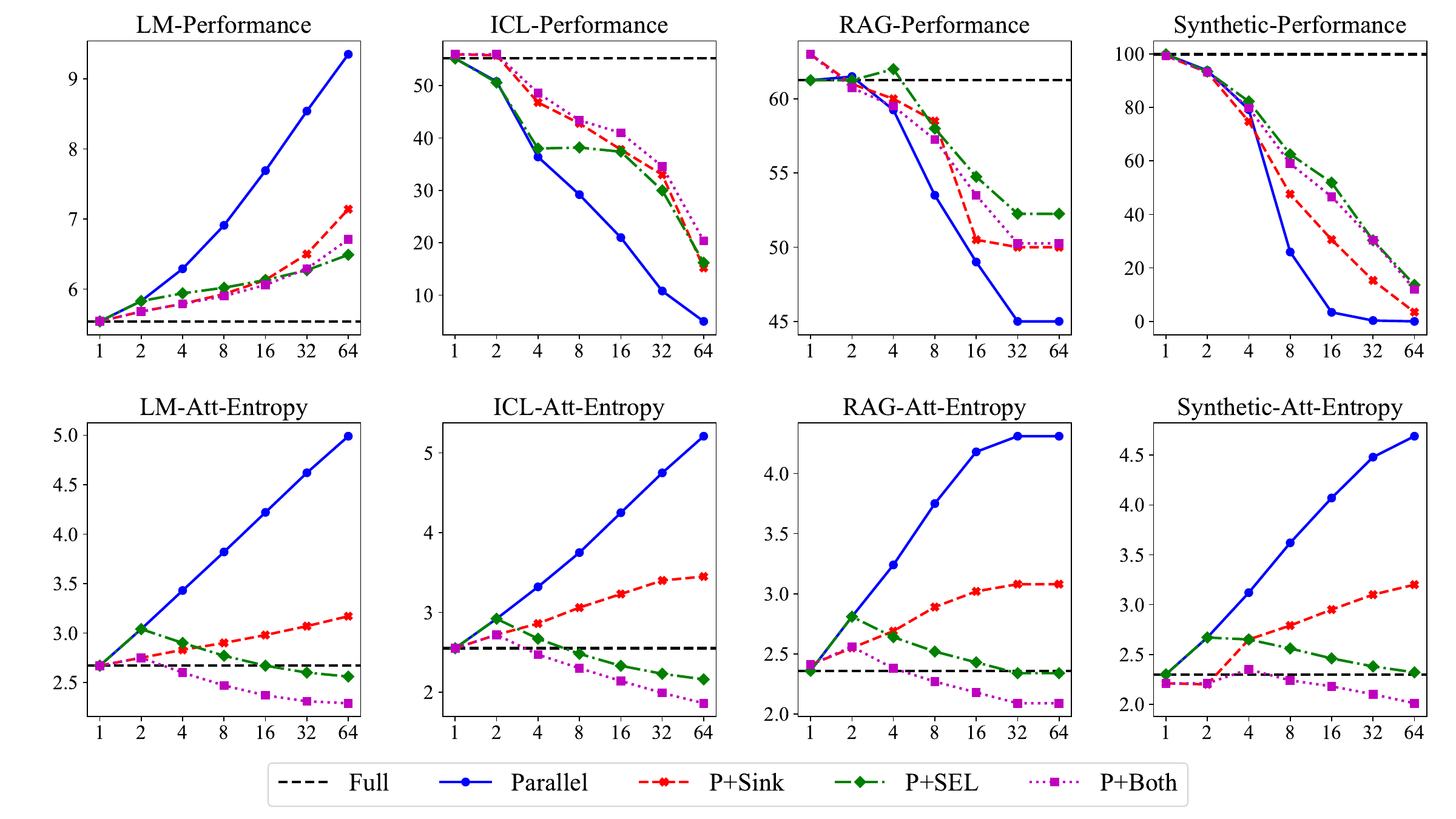}
	\end{subfigure}
	\caption{The influence of the entropy reduction methods (with \textsc{Llama-3.1-8B} and 4K lengths). Notations are the same as those in Table~\ref{fig:methods}.}
	\label{fig:methods3a}
\end{figure*}

\begin{figure*}[t]
	\centering
	\begin{subfigure}[b]{0.95\textwidth}
		\includegraphics[width=0.975\textwidth]{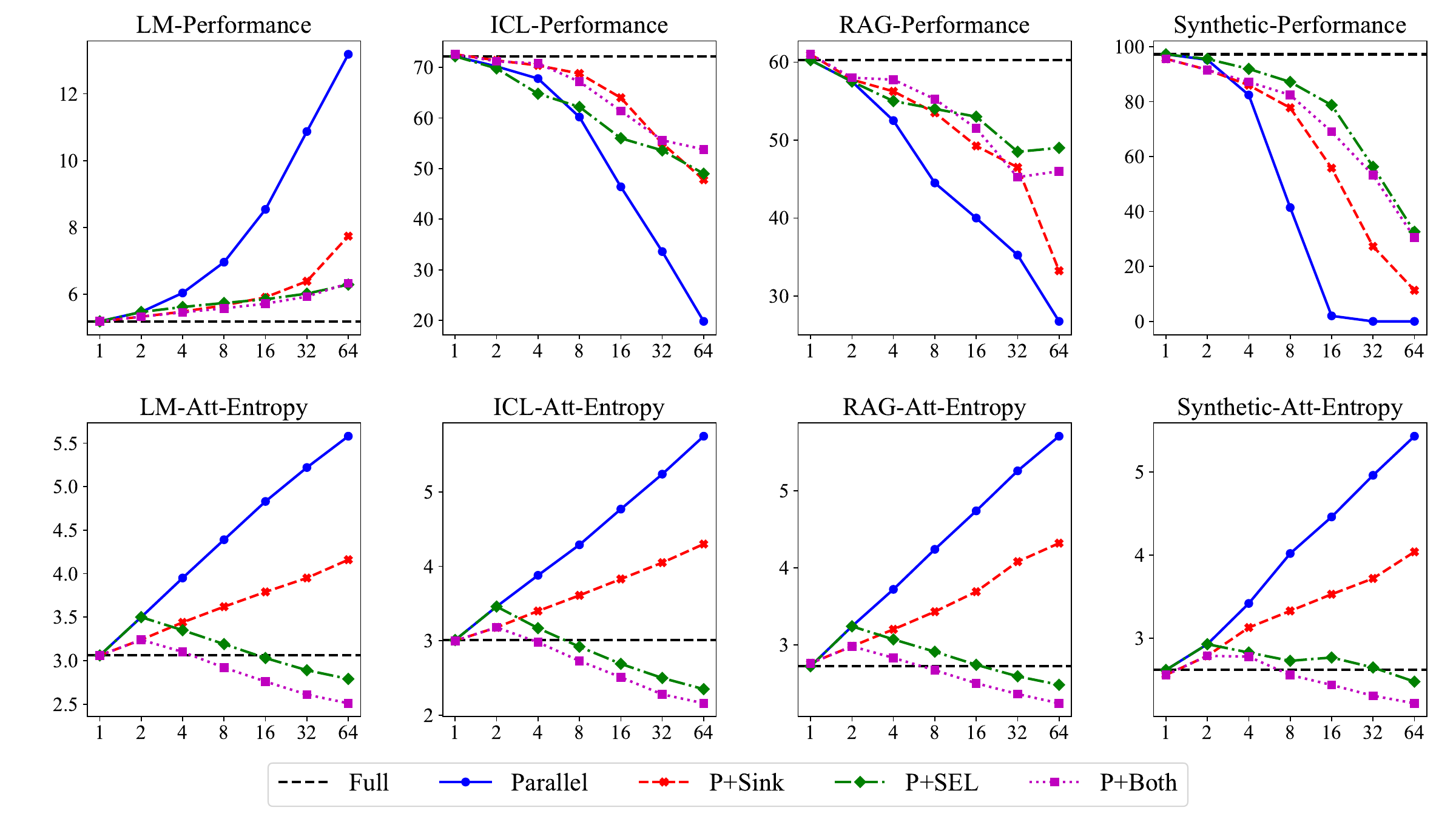}
	\end{subfigure}
	\caption{The influence of the entropy reduction methods (with \textsc{Llama-3.1-8B} and 16K lengths). Notations are the same as those in Table~\ref{fig:methods}.}
	\label{fig:methods3b}
\end{figure*}

\begin{figure*}[t]
	\centering
	\begin{subfigure}[b]{0.95\textwidth}
		\includegraphics[width=0.975\textwidth]{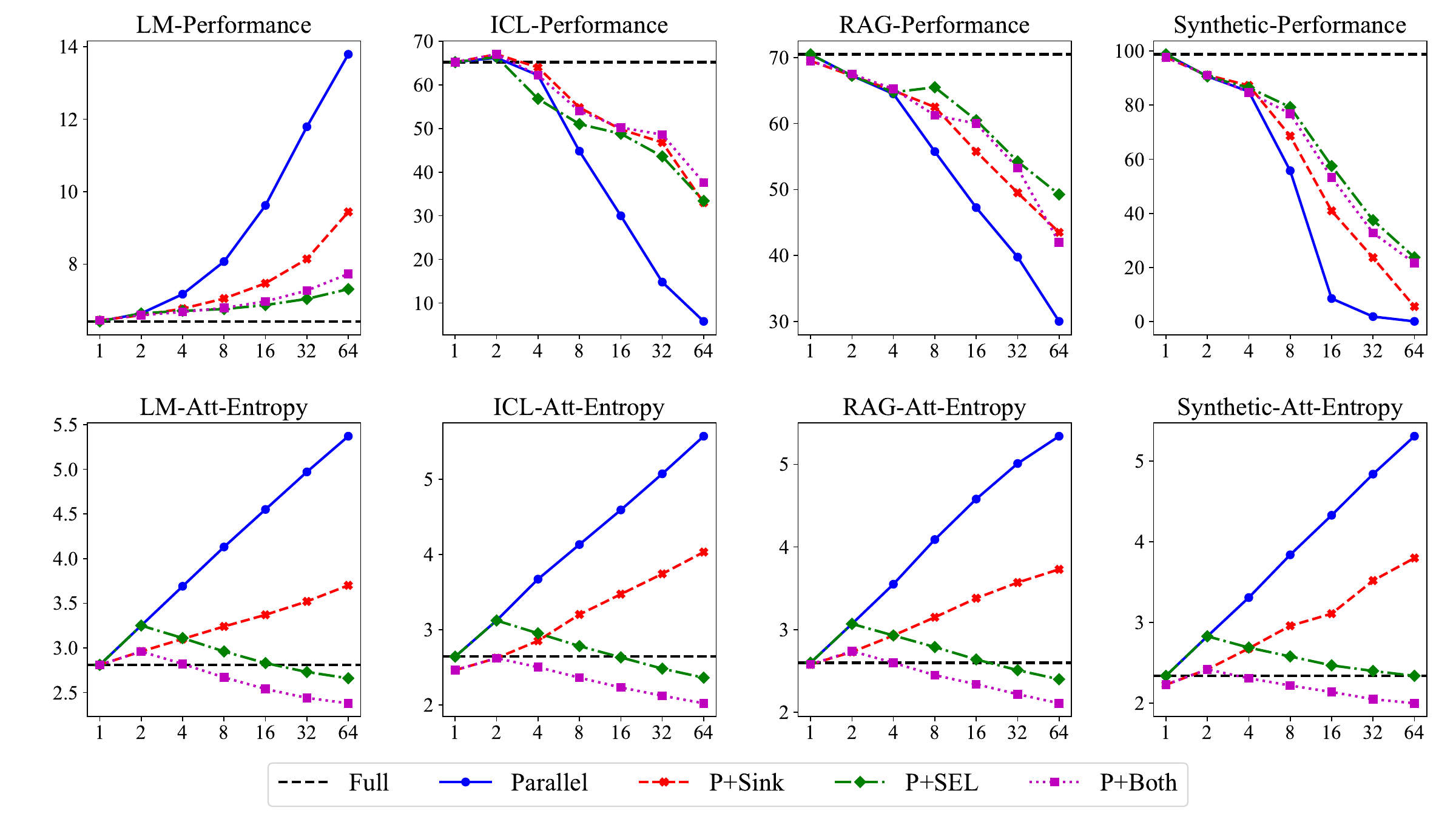}
	\end{subfigure}
	\caption{The influence of the entropy reduction methods (with \textsc{Llama-3.1-8B-Instruct} and 8K lengths). Notations are the same as those in Table~\ref{fig:methods}.}
	\label{fig:methods4}
\end{figure*}

\begin{figure*}[t]
	\centering
	\begin{subfigure}[b]{0.95\textwidth}
		\includegraphics[width=0.975\textwidth]{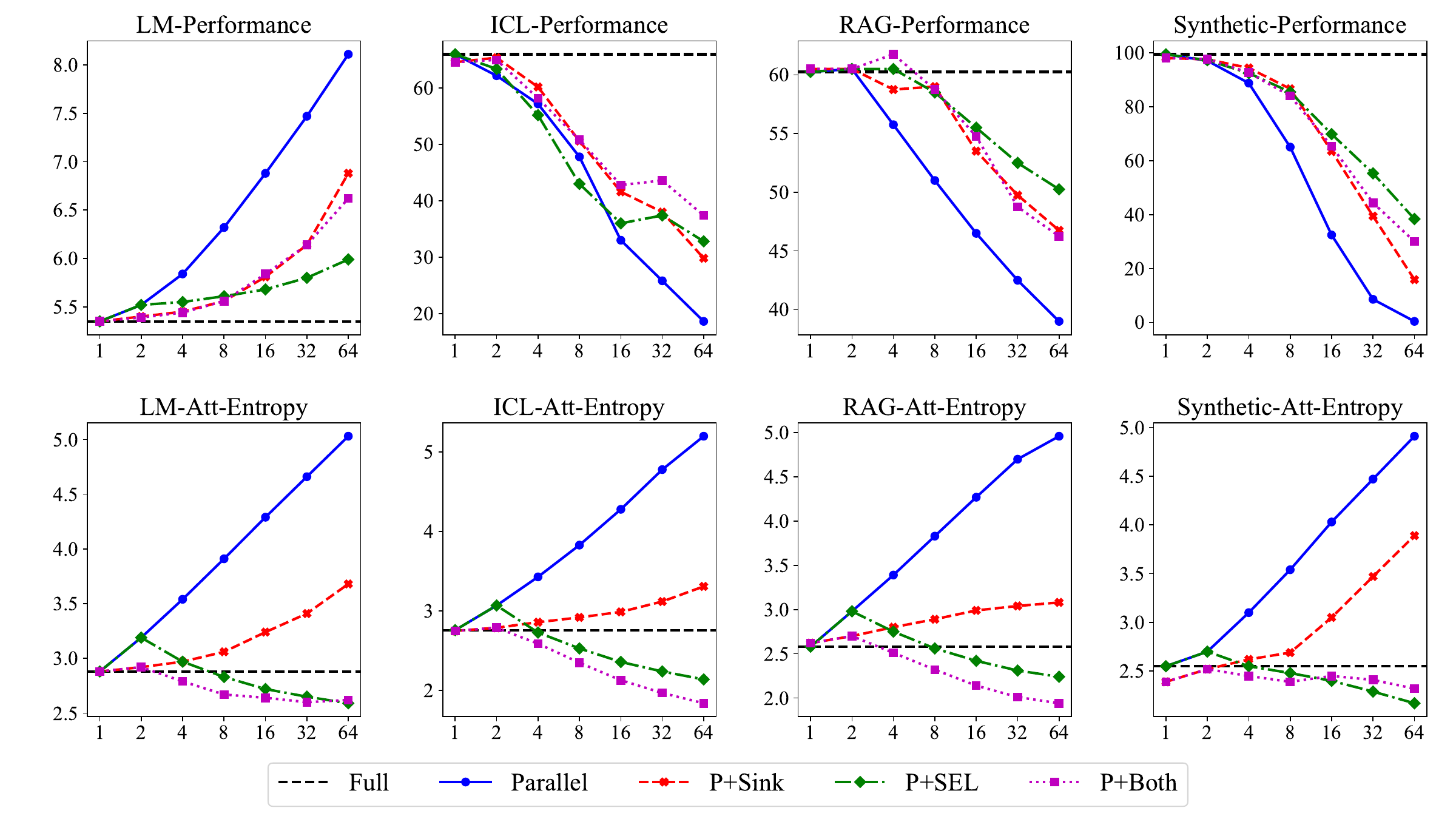}
	\end{subfigure}
	\caption{The influence of the entropy reduction methods using serialized position encoding (with \textsc{Llama-3.1-8B} and 8K lengths). Notations are the same as those in Table~\ref{fig:methods}.}
	\label{fig:methods5}
\end{figure*}

In the appendix, we provide several additional results:
\begin{itemize}
    \item Table~\ref{tab:main_res2} and \ref{tab:main_res3} show the main results of parallel context encoding using \textsc{Mistral-7B-v0.3} and \textsc{Qwen2-7B}, whose patterns are similar to \textsc{Llama-3.1-8B} as shown in Table~\ref{tab:main_res}.
    \item Figure~\ref{fig:corr2} presents the correlations between model performance and attention entropy for other tasks, and the patterns are similar to those in the LM task as shown in Figure~\ref{fig:corr}.
    \item Figure~\ref{fig:methods2}, \ref{fig:methods3}, \ref{fig:methods3a}, \ref{fig:methods3b}, \ref{fig:methods4} and \ref{fig:methods5} illustrate more results of entropy reduction methods in different settings. The overall trends are similar to those in Figure~\ref{fig:methods}.
\end{itemize}

\begin{table}[h]
	\centering
	\small
	\begin{tabular}{l | c c c | c c c | c c c | c c c}
		\toprule
		  & \multicolumn{3}{c|}{LM (PPL$\downarrow$)} & \multicolumn{3}{c|}{ICL (Acc$\uparrow$)} & \multicolumn{3}{c|}{RAG (SubEM$\uparrow$)} & \multicolumn{3}{c}{Synthetic (SubEM$\uparrow$)} \\
        \cmidrule(r){2-4} \cmidrule(r){5-7} \cmidrule(r){8-10} \cmidrule(r){11-13}
		& 4K & 8K & 16K & 4K & 8K & 16K & 4K & 8K & 16K & 4K & 8K & 16K \\
		\midrule
        Full & 5.00 & 4.85 & 4.73 & 48.00 & 55.20 & 68.60 & 56.75 & 55.25 & 56.50 & 98.31 & 97.44 & 89.62 \\
        \midrule
        P=2 & 5.17 & 5.04 & 4.92 & 33.40 & 50.20 & 63.00 & 55.75 & 53.00 & 51.25 & 89.50 & 87.19 & 85.50 \\
        P=4 & 5.43 & 5.33 & 5.26 & 19.20 & 38.20 & 57.00 & 50.50 & 47.25 & 44.50 & 64.69 & 71.19 & 65.69 \\
        P=8 & 5.88 & 5.84 & 5.83 & 11.00 & 23.00 & 44.20 & 44.00 & 40.75 & 39.00 & 16.06 & 16.31 & 18.31 \\
        P=16 & 6.70 & 6.97 & 7.35 & 6.80 & 10.60 & 22.00 & 37.75 & 34.00 & 30.00 & 1.31 & 0.75 & 0.56 \\
        P=32 & 8.55 & 10.09 & 11.96 & 6.00 & 6.60 & 8.00 & 34.00 & 20.75 & 14.25 & 0.38 & 0.06 & 0.12 \\
        P=64 & 11.45 & 15.24 & 20.65 & 3.60 & 4.20 & 6.40 & 34.00 & 8.50 & 2.25 & 0.00 & 0.00 & 0.00 \\
		\bottomrule
	\end{tabular}
	\caption{Comparisons between full-attention and naive parallel encoding with \textsc{Mistral-7B-v0.3}. Notations are the same as those in Table~\ref{tab:main_res}.}
	\label{tab:main_res2}
\end{table}

\begin{table}[h]
	\centering
	\small
	\begin{tabular}{l | c c c | c c c | c c c | c c c}
		\toprule
		  & \multicolumn{3}{c|}{LM (PPL$\downarrow$)} & \multicolumn{3}{c|}{ICL (Acc$\uparrow$)} & \multicolumn{3}{c|}{RAG (SubEM$\uparrow$)} & \multicolumn{3}{c}{Synthetic (SubEM$\uparrow$)} \\
        \cmidrule(r){2-4} \cmidrule(r){5-7} \cmidrule(r){8-10} \cmidrule(r){11-13}
		& 4K & 8K & 16K & 4K & 8K & 16K & 4K & 8K & 16K & 4K & 8K & 16K \\
		\midrule
        Full & 7.33 & 7.27 & 7.01 & 29.60 & 42.80 & 53.20 & 63.50 & 66.00 & 60.75 & 68.44 & 60.69 & 67.50 \\
        \midrule
        P=2 & 7.53 & 7.50 & 7.25 & 34.20 & 35.20 & 51.20 & 62.00 & 57.50 & 57.00 & 67.88 & 41.06 & 37.06 \\
        P=4 & 8.15 & 7.72 & 7.71 & 23.20 & 34.60 & 44.40 & 58.25 & 54.25 & 51.00 & 25.38 & 45.00 & 13.88 \\
        P=8 & 8.51 & 8.97 & 8.13 & 17.80 & 28.20 & 45.60 & 53.00 & 48.25 & 48.50 & 4.12 & 3.56 & 14.06 \\
        P=16 & 9.23 & 9.62 & 10.56 & 12.20 & 17.40 & 31.00 & 46.25 & 39.75 & 38.75 & 0.31 & 0.50 & 1.25 \\
        P=32 & 10.40 & 10.92 & 11.96 & 6.40 & 10.20 & 21.40 & 42.75 & 38.00 & 33.25 & 0.00 & 0.00 & 0.00 \\
        P=64 & 12.12 & 13.09 & 14.67 & 4.00 & 5.60 & 11.40 & 42.75 & 29.75 & 25.50 & 0.00 & 0.00 & 0.00 \\
		\bottomrule
	\end{tabular}
	\caption{Comparisons between full-attention and naive parallel encoding with \textsc{Qwen2-7B}. Notations are the same as those in Table~\ref{tab:main_res}.}
	\label{tab:main_res3}
\end{table}

\end{document}